\documentclass[runningheads]{llncs}

\usepackage{tikz}
\usepackage[paperheight=235mm,
paperwidth=155mm,textwidth=12.2cm,textheight=19.3cm]{geometry}
\usepackage{bbding}

\usepackage[ruled,vlined,linesnumbered]{algorithm2e}
\usepackage{amsmath}
\usepackage{wrapfig}

\usepackage{tikz,pgfplots}
\pgfplotsset{compat=1.11}
\usepgfplotslibrary{groupplots}
\usetikzlibrary{decorations}
\usetikzlibrary{decorations.markings}
\usepgfplotslibrary{fillbetween}
\pgfplotsset{%
	actfunc/.style = {%
		domain=-5:3,
		samples = 400,
		smooth,
		thick,
		on layer={axis foreground},
	}
}

\usepackage{bigints}
\usepackage{mathtools}
\usepackage{multirow}

\definecolor{mygreen}{rgb}{0,0.6,0}
\definecolor{mygray}{rgb}{0.5,0.5,0.5}
\definecolor{mymauve}{rgb}{0.58,0,0.82}
\definecolor{dkgreen}{rgb}{0,0.5,0}

\definecolor{lightgray}{rgb}{0.85,0.85,0.85}
\definecolor{lightgreen}{rgb}{0.7,0.9,0.7}
\definecolor{lightblue}{rgb}{0.7,0.7,0.9}
\definecolor{lightred}{rgb}{0.9,0.7,0.7}

\newcommand{\ignore}[1]{}

\usepackage{graphicx}
\usepackage{listings}
\usepackage{mathtools}
\usepackage{xcolor}

\usepackage{graphicx}
\usepackage{amsmath}
\usepackage{amssymb}
\usepackage{booktabs}
\usepackage{comment}

\usepackage{tikz}
\usetikzlibrary{positioning}
\usetikzlibrary{backgrounds}
\usetikzlibrary{calc}
\usetikzlibrary{fit}
\usetikzlibrary{positioning}
\usetikzlibrary{shadows}
\usetikzlibrary{shapes.geometric}
\usetikzlibrary{shapes.multipart}

\definecolor{champagne}{rgb}{0.97, 0.97, 0.81}

\definecolor{maincolor}{rgb}{0.1,0.5,1}
\colorlet{lightmain}{maincolor!75}
\colorlet{lightermain}{maincolor!50}
\colorlet{lightestmain}{maincolor!25}
\colorlet{darkmain}{maincolor!75!black}
\colorlet{darkermain}{maincolor!50!black}
\colorlet{darkestmain}{maincolor!25!black}

\newcommand{\Name}{\textsc{LinSyn}}

\newcommand{\Popqorn}{\textsc{Popqorn}}

\newcommand{\dReal}{\textsc{dReal}}
\newcommand{\autolipra}{\textsc{AutoLiRPA}}
\newcommand{\popqorn}{\textsc{POPQORN}}

\begin{document}

\title{\Name{}: Synthesizing Tight Linear Bounds for Arbitrary Neural Network
Activation Functions\thanks{This work was partially funded by the U.S. National
Science Foundation
	grants CNS-1813117 and CNS-1722710, and the U.S. Office of Naval
	Research (ONR) grant N00014-17-1-2896.}}

\author{Brandon Paulsen\Envelope \and Chao Wang}

\authorrunning{}

\institute{
University of Southern California, Los Angeles CA 90089, USA\\
\email{\{bpaulsen,wang626\}@usc.edu}
}

\maketitle

\begin{abstract}
	The most scalable approaches to certifying neural network robustness depend
	on computing sound linear lower and upper bounds for the network's
	activation functions.
	Current approaches are limited in that the linear bounds must be
	handcrafted by an expert, and can be sub-optimal, especially when the
	network's architecture composes operations using, for example,
	multiplication such as in LSTMs and the recently popular \emph{Swish} activation.
	The dependence on an expert prevents the application of robustness
	certification to developments in the state-of-the-art of activation
	functions, and furthermore the lack of tightness guarantees may give a
	false sense of insecurity about a particular model.
	%
	To the best of our knowledge, we are the first to consider the problem of
	\textit{automatically} synthesizing \textit{tight} linear bounds for
	arbitrary n-dimensional activation functions.
	We propose the first fully automated method that achieves tight linear bounds
        while only leveraging the
	mathematical definition of the activation function itself.
	Our method leverages an efficient heuristic technique to synthesize bounds
	that are tight and \emph{usually sound}, and then
	verifies the soundness (and adjusts the bounds if necessary) using the
	highly optimized branch-and-bound SMT solver, \dReal{}.
	Even though our method depends on an SMT solver, we show that the runtime
	is reasonable in practice, and, compared with state of the art, our
	method often achieves 2-5X tighter final output bounds and more than
	quadruple certified robustness.



\end{abstract}

\section{Introduction}

Prior work has shown that neural networks are vulnerable to various types of
(adversarial) perturbations, such as small $l$-norm bounded
perturbations~\cite{szegedy2013intriguing}, geometric
transformations~\cite{engstrom2019exploring,kanbak2018geometric}, and word
substitutions~\cite{alzantot2018generating}. Such perturbations can often
cause a misclassification for any given input, which may have serious
consequences, especially in safety critical systems.
Certifying robustness to these perturbations has become an important problem as
it can show the network does not exhibit these misclassifications, and
furthermore
previous work has shown that a given input feature's
certified robustness can be a useful indicator to determine the feature's
importance in the network's decision~\cite{shi2020robustness,ko2019popqorn}.

Indeed, many approaches have been proposed for certifying the
robustness of inputs to these perturbations. Previous work typically
leverages two types of techniques: (1) fast and scalable, but approximate
techniques~\cite{SinghGPV19,GehrMDTCV18,WengZCSHDBD18,shi2020robustness,ko2019popqorn},
 and (2) expensive but exact techniques that leverage some type of
constraint
solver~\cite{KatzBDJK17,KatzHIJLLSTWZDK19,tjeng2019evaluating}. Several works
have also
combined the
two~\cite{SinghGPV19iclr,Singh2019krelu,WangPWYJ18nips,tran2019star}. The
most successful approaches, in terms of scalability in practice, are built on top of the approximate techniques, which
all depend on computing \textit{linear bounds} for the non-linear activation functions.



However, a key limitation is that the linear bounds must be handcrafted and
proven sound by experts. Not only is this process difficult, but also ensuring
the tightness of the crafted bounds presents an additional challenge.
Unfortunately, prior work has only crafted bounds for the most common
activation functions and architectures, namely ReLU~\cite{WangPWYJ18nips},
sigmoid, tanh~\cite{SinghGPV19,zhang2018efficient,wu2021tightening}, the exp
function~\cite{shi2020robustness}, and some
2-dimensional activations found in LSTM networks~\cite{ko2019popqorn}.
As a result, existing tools for neural network verification
cannot handle a large number of activation functions that are
frequently used in practice.  Examples include the \emph{GELU}
function~\cite{hendrycks2016gaussian}, which is currently the activation
function used in OpenAI's GPT~\cite{radford2018improving}, and the \emph{Swish}
function which has been shown to outperform the standard ReLU function in some
applications~\cite{ramachandran2017searching} and, in particular, can reduce
over-fitting in adversarial training~\cite{singla2021low}. In addition, these
recently introduced activation
functions are often significantly more complex than previous activation
functions, e.g., we have $
\mathit{gelu}(x) = 0.5 x ( 1 + \tanh{[ \sqrt{2 / \pi } (x + 0.044715x ^{3} ) ]
} )
$.
%



In this work, we study the problem of \textit{efficiently} and
\textit{automatically}
synthesizing \textit{sound} and \textit{tight} linear bounds for any
\textit{arbitrary activation function}.
By \emph{arbitrary activation function}, we mean \textit{any} (non-linear)
computable function $ z = \sigma(x_1,\dots,x_d) $ used inside a neural network
with $ d $ input variables.
By \textit{sound} we mean, given an interval bound on each
variable $ x_1 \in
[l_1, u_1], x_2 \in [l_2, u_2], \dots, x_d \in [l_d, u_d]$, the problem is to \textit{efficiently}
compute lower bound coefficients $ c^l_1, c^l_2, \dots, c^l_{d+1} $, and upper
bound coefficients $ c^u_1, c^u_2, \dots, c^u_{d+1} $ such that the following
holds:
\begin{equation}\label{eq:intro-sound}
\begin{gathered}
	\forall x_1 \in [l_1, u_1], x_2 \in [l_2, u_2], \dots,  x_d \in [l_d, u_d]\\
	c^l_1x_1 + c^l_2x_2 + \dots + c^l_{d+1}
	\leq \sigma(x_1,\dots,x_d) \leq
	c^u_1x_1 + c^u_2x_2 + \dots + c^u_{d+1}
\end{gathered}
\end{equation}
By \textit{automatically}, we mean that the above is done using only the
definition of the activation function itself.
Finally, by \textit{tight}, we mean that some formal measure, such as the
volume above/below the linear bound, is minimized/maximized.

\begin{figure}[t]
	\centering
	\includegraphics[width=.9\linewidth]{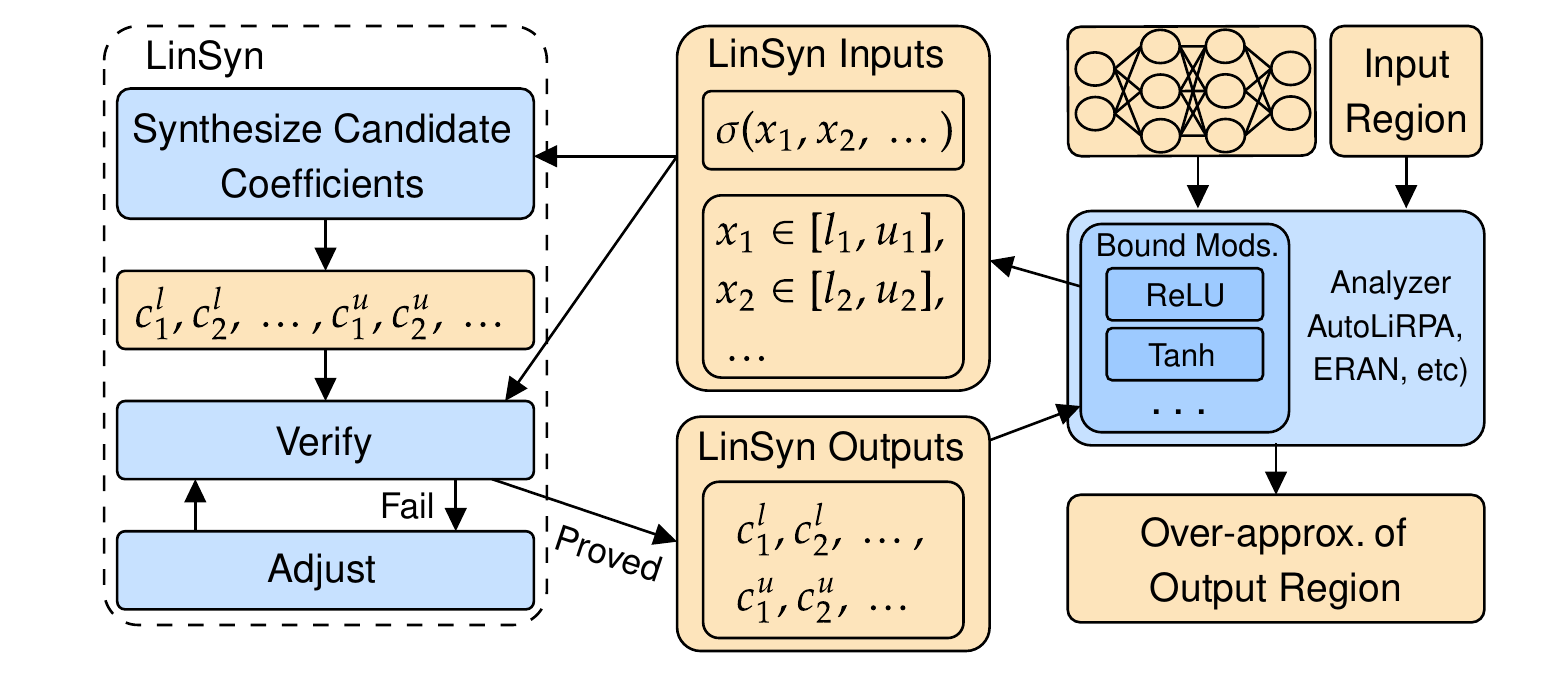}
	\caption{The overall flow of~\Name{}.\label{fig:block-diagram}}
\end{figure}


We have developed a new method, named~\Name{}, that can \emph{automatically}
synthesize tight linear bounds for \emph{any arbitrary} non-linear activation
function $ \sigma(\cdot) $.
We illustrate the flow of our method on the left-hand side of
Fig.~\ref{fig:block-diagram}. As shown,~\Name{} takes two inputs: a
definition of the activation function, and an interval for each of its
inputs.~\Name{} outputs linear coefficients such that
Equation~\ref{eq:intro-sound} holds.
Internally,~\Name{}
uses sampling and an LP (linear programming) solver to synthesize
candidate lower and upper bound coefficients.
Next, it uses an
efficient local minimizer to compute a good estimate of the offset
needed to ensure soundness of the linear bounds.
Since the candidate bounding functions constructed in this manner may
still be unsound, finally, we use a highly optimized branch-and-bound
nonlinear SMT solver, named~\dReal{}~\cite{gao2013dreal}, to verify the
soundness of the linear bounds.
Even though our new method involves the use of solvers and
optimizers, the entire process typically takes less than 1/100th of a
second per pair of bounds.

Fig.~\ref{fig:block-diagram} also illustrates how~\Name{} fits in with
existing neural network verification frameworks, such as ERAN~\cite{eran},
and~\autolipra{}~\cite{autolipra}. These tools take as input a neural network,
and a region of the neural networks input space, and compute an
over-approximation of the neural network's outputs. Internally, these
frameworks have modules that compute linear bounds for a specific activation
functions.~\Name{} is a one-size-fits-all drop-in replacement for these modules
that are invoked at runtime whenever a linear bound of a non-linear activation function is needed.





Our method differs from these existing frameworks because a user (usually an
expert in neural network verification) must provide
hand-crafted, sound linear bounds for the activation functions of a
neural network.
%
%
However, to date, they only support the previously mentioned activation
functions. We note however that the recent framework~\autolipra{} supports
binary operations (namely addition, subtraction, multiplication, and division)
as ``activation functions''. Thus, while it's not explicitly designed to handle
complex
activations, it has the ability to by decomposing, e.g., $ gelu(x) $ into
operations that it supports, and then combining them. In contrast,~\Name{}
bounds the activation function \textit{as a whole}, which we will show produces
much tighter linear bounds.
%




We have implemented our method in tool called \Name{}, and evaluated it on
benchmarks in computer vision and natural language processing (NLP).
Our evaluation shows that we can obtain final output bounds often 2-5X
tighter than the most general tool~\cite{autolipra}, thus allowing us to drastically
increase certified robustness.
In addition, our tool achieves accuracy equal to or better
than the handcrafted LSTM bounds of \Popqorn{}~\cite{ko2019popqorn}, which is
currently the most accurate tool for analyzing LSTM-based NLP models, at a
comparable runtime.


To summarize, this paper makes the following contributions:
\begin{itemize}
	\item We propose the first method for automatically synthesizing  tight linear bounds for
	arbitrary activation functions.
	\item We implement our approach in a tool called~\Name{}, and integrate it
	as a bounding module into the~\autolipra{} framework, thus producing a
	neural network verification tool that can theoretically compute tight linear
	bounds for any arbitrary activation function.
	\item We extensively evaluate our approach and show it outperforms
	state-of-the-art tools in terms of accuracy and certified robustness by a
	large margin.
\end{itemize}

The rest of this paper is organized as follows.  First, we provide the
technical background in Section~\ref{sec:preliminaries}.  Then, we
present our method for synthesizing the linear bounds in
Section~\ref{sec:method-1} and our method for verifying the linear
bounds in Section~\ref{sec:method-2}.  Next, we present the
experimental results in Section~\ref{sec:experiment}.  We review the
related work in Section~\ref{sec:related} and, finally, give our
conclusions in Section~\ref{sec:conclusion}.

\section{Preliminaries}
\label{sec:preliminaries}

In this section, we define the neural network verification problem, and
illustrate both how state-of-the-art verification techniques work, and their
limitations.

\subsection{Neural Networks}


Following conventional notation, we refer to matrices with capital bold letters
(e.g. $ \mathbf{W} \in \mathbb{R}^{n \times m}$), vectors as lower case bold
letters (e.g. $ \mathbf{x} \in \mathbb{R}^n $), and scalars or variables with
lower case letters (e.g. $ x \in \mathbb{R} $). Slightly deviating from the
convention, we refer to a set of elements with capital letters (e.g. $ X
\subseteq \mathbb{R}^n $).

We consider two types of networks in our work: feed-forward and recurrent. We
consider a feed-forward neural network to be a (highly) non-linear function $f :
\mathbb{X} \to \mathbb{Y} $, where $\mathbb{X} \subseteq \mathbb{R}^n$ and $
\mathbb{Y} \subseteq \mathbb{R}^m $. We focus on neural network
\textit{classifiers}. For an input $ \mathbf{x} \in \mathbb{X} $, each element
in the output $ f(\mathbf{x}) $ represents a score for a particular class, and
the class
associated with the largest element is the chosen class. For example, in image
classification, $ \mathbb{X} $ would be the set of all images, each element of
an input $ \mathbf{x} \in \mathbb{X} $ represents a pixel's value, and each
element in $ \mathbb{Y} $ is associated with a particular object that the image
might contain.

In feed-forward neural networks the output $ f(\mathbf{x}) $ is computed by
performing a
series of affine transformations, i.e., multiplying by a weight matrix, followed
by application of an activation function $ \sigma(\cdot) $. Formally, a neural
network with $ l $ layers has $ l $ two-dimensional weight matrices and $ l $
one-dimensional bias vectors $ \mathbf{W_i}, \mathbf{b_i},$ where $i
\in {1..l}
$, and thus we have $ f(\mathbf{x}) = \mathbf{W_l} \cdot \sigma(\mathbf{W_{l-1}}
\dots \cdot \sigma(\mathbf{W_1} \cdot \mathbf{x} + \mathbf{b_1}) \dots +
\mathbf{b_{l-1}}) + \mathbf{b_l} $, where $ \sigma(
\cdot ) $
is the activation function applied element-wise
to the input vector. The default choice of activation is typically the
sigmoid $ \sigma(x) = 1 / (1 + e^{-x}) $, $ \tanh{} $, or ReLU
function $\sigma(x) = max(0, x) $, however recent
work~\cite{hendrycks2016gaussian,ramachandran2017searching,radford2018improving}
 has shown that functions such as $ gelu(x) $ and $ \mathit{swish}(x) = x
\times \mathit{sigmoid}(x) $ can have better performance and desirable
theoretical properties.

Unlike feed-forward neural networks, recurrent neural networks receive a
sequence of inputs $ [\mathbf{x^{(1)}}, \dots, \mathbf{x^{(t)}} ] $, and the
final
output of $ f $ on $ \mathbf{x_t} $ is used to perform the classification of
the whole sequence. Recurrent neural networks are
\textit{state-ful}, meaning they maintain a state vector
that contains information about inputs previously given to
$ f $, which also gets updated on each call to $ f $.
In particular, we focus on \textit{long short-term memory} (LSTM) networks, which
have seen wide adoption in natural language processing (NLP) tasks due to their
sequential nature. For LSTMs trained for NLP tasks, the network receives a
sequence of \textit{word embeddings}. A word embedding is an $ n $-dimensional
vector that is associated with a particular word in a (natural) language. The
distance between word embeddings carries semantic significance -- two word
embeddings that are close to each other in $ \mathbb{R}^n $ typically have
similar meanings or carry a semantic relatedness (e.g. \textit{dog} and
\textit{cat} or \textit{king} and \textit{queen}), whereas unrelated words
typically are farther apart.

LSTM networks further differ from feed-forward networks in
that their internal activation functions are \textit{two}-dimensional.
Specifically, we have the following two activation patterns: $ \sigma_1(x)
\times \sigma_2(y) $ and $ x \times \sigma_1(y) $. The default choices are $
\sigma_1(x) = sigmoid(x) $, and $ \sigma_2(x) = tanh(x) $. However, we can swap
$ \sigma_1 $ with any function with output range bounded by $ [0, 1] $, and
swap $ \sigma_2 $ with any function with output range bounded by $ [-1, 1] $.
Indeed, prior work~\cite{gomes2008complementary} has shown that $ \sigma_1(x) =
1 - e^{e^{-x}} $ can achieve better results in some applications.


%

\subsection{Neural Network Verification}



A large number of problems in neural network verification can be phrased as the
following:
given an input region $ X \subseteq \mathbb{X} $, compute an
over-approximation $ Y $, such that $ \{ f(\mathbf{x})\;  | \; \mathbf{x} \in X
\} \subseteq Y \subseteq \mathbb{Y} $. Typically $ X $ and $ Y
$ are hyper-boxes represented by an interval for each of their elements.
A common problem is to prove that a point $ \mathbf{x} \in
\mathbb{X} $ is \textit{robust}, meaning that small perturbations will not
cause an incorrect classification. In this case, $ X $
is the set of all perturbed versions of $ \mathbf{x} $, and
to prove robustness, we check that the element of the correct class in $ Y $
has a lower bound that is greater than the upper bound of all other elements.

\begin{figure}[t]
	\centering
	\begin{minipage}{.63\textwidth}
		\centering
		\includegraphics[width=\linewidth]{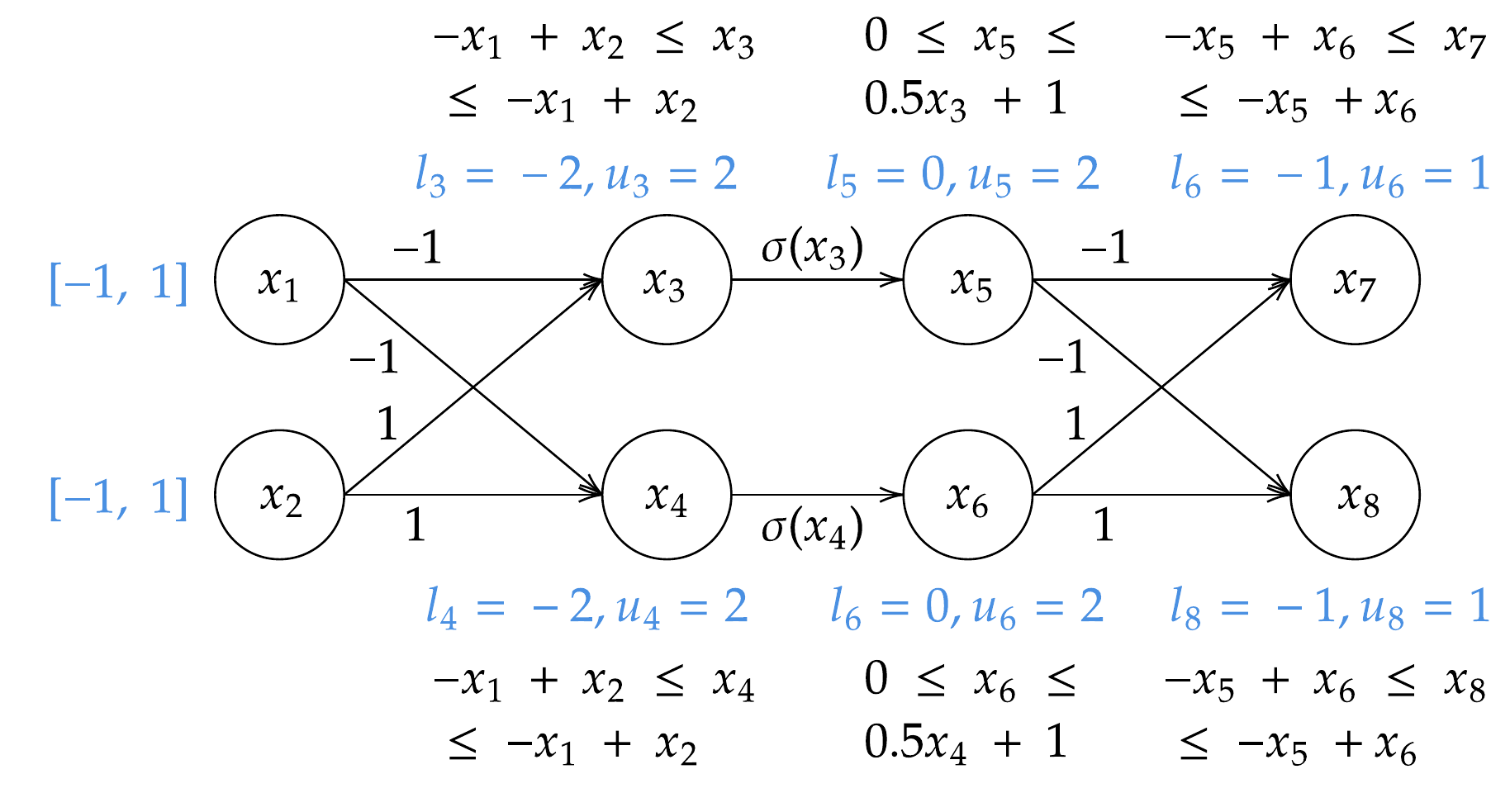}
		\caption{Example of neural network verification.}
		\label{fig:motex}
	\end{minipage}\hspace{24pt}%
	\begin{minipage}{.27\textwidth}
		\centering
		\includegraphics[width=\linewidth]{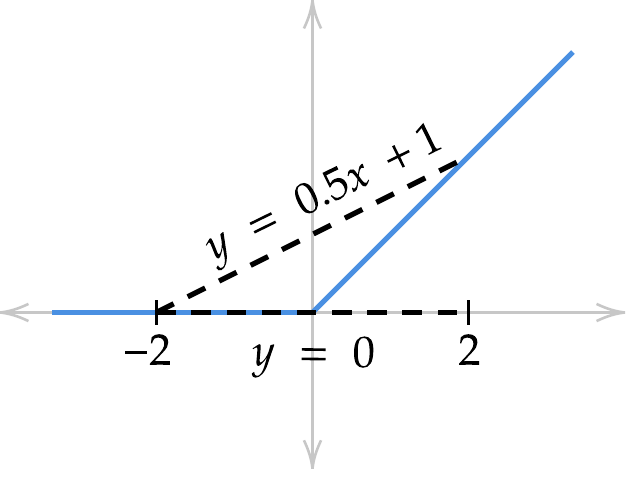}
		\caption{Linear bounds for ReLU activation.}
		\label{fig:linearbound}
	\end{minipage}
\end{figure}

We illustrate a simple verification problem on the neural network shown in
Fig.~\ref{fig:motex}. The network has two inputs, $ x_1, x_2 $, and two
outputs
$ x_7, x_8 $ which represent scores for two different classes. We refer to the
remaining hidden neurons as $ x_i, i \in {3..6} $. Following prior
work~\cite{SinghGPV19}, we break the affine transformation and application of
the activation function into two separate neurons, and the neurons are assumed
to be ordered such that, if $ x_i $ is in a layer before $ x_j $, then $ i < j
$.
For simplicity, in this motivating example, we let $ \sigma(x) = max(0, x) $ (the ReLU function).
We are interested in proving that the region $ x_1 \in [-1, 1], x_2 \in [-1 ,1]
$ always maps to the first class, or in other words, we want to show that the
lower bound of $ x_7 $ is greater than the upper bound $ x_8 $.

\subsection{Existing Methods}

The most scalable approaches (to date) for neural network verification are
based on linear bounding and back-substitution~\cite{autolipra}, also referred
to
as abstract interpretation in the polyhedral abstract domain~\cite{SinghGPV19} or symbolic
interval analysis~\cite{WangPWYJ18nips} in prior work.

For each
neuron $ x_j $ in the network, these approaches compute a concrete lower and
upper bound $ l_j, u_j $, and a linear lower and upper bound in terms of the
previous layer's neurons. The linear bounds (regardless of the choice of $
\sigma(\cdot) $) have the following form:
$ \sum_{i=0}^{j-1} x_i \cdot c^l_i + c^l_j \leq x_j \leq \sum_{i=0}^{j-1}
x_i \cdot c^u_i + c^u_j $. The bounds are computed in a forward, layer-by-layer
fashion which guarantees that any referenced neurons will already have a bound
computed when back-substitution is performed.

To obtain the concrete bounds $ l_j, u_j $ for a neuron $ x_j $, the bounds of
any non-input neurons are recursively substituted into the linear bounds of $
x_j $ until only input nodes $ x_1, ..., x_n $ remain. Finally, the concrete
input intervals are substituted into the bound to obtain $ l_j, u_j $.

\paragraph{Example}

We illustrate on the two-layer network in Fig.~\ref{fig:motex} for the
previously defined property. We trivially have $ l_1 = l_2 = -1 $, $ u_1 = u_2
= 1 $, $ -1 \leq x_1 \leq 1 $, and $ -1 \leq x_2 \leq 1 $. We then compute
linear bounds for $ x_3, x_4 $ in terms of previous layer's neurons $ x_1, x_2
$.
We multiply $ x_1, x_2 $ by the edge weights, obtaining $ -x_1 + x_2 $ as the
lower and upper bound for both of $ x_3 $ and $ x_4 $.
Since this bound is already in terms of the input variables, we substitute the
concrete bounds into this equation and obtain $ l_3 = l_4 = -2 $ and $ u_3 =
u_4 = 2 $.

Next, we need to compute the linear bounds for $ x_5 = \sigma(x_3) $ and $ x_6
= \sigma(x_4) $ after
applying the activation function. Solving this challenge has been the focus of
many prior works. There are two requirements. First, they need to be
\textit{sound}. For example, for $ x_5 $ we need to find coefficients $
c_1^l,c_2^l,c_1^u,c_2^u $ such that
$c_1^l x_3 + c_2^l
\leq \sigma(x_3) \leq c_1^ux_3 + c_2^u $ for all $x_3 \in [l_3, u_3]$, and similarly for $ x_6 $. Second, we
want them to be \textit{tight}. Generally, this means that volume below the
upper bound is minimized, and volume below the lower bound is maximized.

As an example, prior work~\cite{SinghGPV19,zhang2018efficient} proposed the following sound and
tight bound for $ \sigma(x) = max(0, x) $:
\[
	\forall x_i \in [l_i, u_i] ~.~ \frac{u_i}{u_i - l_i}x_i +
	\frac{-l_iu_i}{u_i-l_i} \leq \sigma(x_i)
	\leq \begin{cases}
	0 & -l_i \geq u_i\\
	x_i & -l_i < u_i
	\end{cases}
\]
We illustrate the bound for $ x_5 $ in Fig.~\ref{fig:linearbound}. After
computing this bound, we recursively substitute variables in the
bounds of $ x_5 $ with the appropriate bound, and compute $ l_5, u_5 $. The
process then repeats for $ x_6 $, followed by $ x_7 $ and $ x_8 $. We then
check $ l_7 > u_8 $ to verify the property, which fails in this case.

\subsection{Limitations of Existing Methods}

Current approaches only support a limited number of activation functions, and
designing linear bounds for new activation functions often requires a
significant amount of effort even for a domain expert.
%
For example, handcrafted sound and
tight linear
bounds for activation functions such as ReLU, sigmoid, and
tanh~\cite{SinghGPV19,WengZCSHDBD18,zhang2018efficient,wu2021tightening,WangPWYJ18,WangPWYJ18nips},
convolution layers and pooling operations~\cite{boopathy2019cnn}, the
two-dimensional activations found in
LSTMs~\cite{ko2019popqorn,ryou2021scalable}, and those
 in transformer networks~\cite{shi2020robustness} are worthy of publication.
Furthermore, even bounds that are hand-crafted by experts are not always
tight. For example, a recent
work~\cite{wu2021tightening} was able to nearly triple the precision
of previous state-of-the-art sigmoid and tanh linear bounds simply by
improving tightness.


To the best of our knowledge,~\autolipra{}~\cite{autolipra} is the only tool
that has the ability to handle more complex activation functions, though it was
not originally designed for this. It
can do so by decomposing them into simpler
operations, and then composing the bounds together. We illustrate with $
swish(x) = x \times sigmoid(x) $, where $ x \in [-1.5, 5.5] $.~\autolipra{}
would first bound $ sigmoid(x) $ over the region $ [-1.5, 5.5] $, resulting in
the bound $ .11x + .35 \leq sigmoid(x) \leq .22x + .51 $.
For the left-hand side of the function, we
trivially have $ x \leq x \leq x $.
~\autolipra{} would then bound a multiplication $ y \times z $, where
in this case $ y = x $ and $ z = sigmoid(x) $, resulting in the final bound
$ -.15x - .495 \leq x\times sigmoid(x) \leq 0.825x + .96 $. We illustrate this
bound in Fig.~\ref{fig:boundcompare2}, and we provide bounds computed
by~\Name{} as a comparison point.~\Name{} provides a slightly better upper
bound, and a significantly better lower bound. The reason for the looseness is
because when~\autolipra{} bounds $ sigmoid(x) $, it necessarily accumulates some
approximation error because it is approximating the behavior of a non-linear
function with linear bounds. The approximation error effectively ``loses some
information'' about about its input variable $ x $. Then, when bounding the
multiplication operation, it has partially lost the information that $ y $ and
$ z $ are related (i.e. they are both derived from $ x $).
In
contrast,~\Name{} overcomes this issue by considering $ swish(x) $ as a whole.
We explain how in the following sections.


\section{Synthesizing the Candidate Linear Bounds}
\label{sec:method-1}

In this section, we describe our method for synthesizing candidate, possibly
unsound linear bounds.

\subsection{Problem Statement and Challenges}


We assume we are given a $ d $-dimensional activation function $ z =
\sigma(x_1,...,x_d) $, and an input interval $ x_i \in [l_i, u_i] $ for each $ i
\in \{1..d\} $. Our goal is to synthesize linear coefficients $ c^l_i, c^u_i$, where $i
\in \{1..d+1\} $ that are sound, meaning that the following condition holds:
\begin{equation} \label{eq:generalsound}
\begin{gathered}
\forall x_1 \in [l_1, u_1], x_2 \in [l_2, u_2], \dots, x_d \in [l_d, u_d]\\
c^l_1x_1 + c^l_2x_2 + \dots + c^l_{d+1}
\leq \sigma(x_1, x_2, \dots) \leq
c^u_1x_1 + c^u_2x_2 + \dots + c^u_{d+1}
\end{gathered}
\end{equation}

In addition, we want to ensure that the bounds are \textit{tight}. The ideal
definition of tightness would choose linear bounds that maximize the precision
of the overall analysis, for example minimizing the width of the output
neuron's intervals.
Unfortunately, such a measure would involve all of the neurons of the network,
and so is impractical to compute.
Instead, the common practice is to settle for tightness that's local to the specific neuron we are
bounding.

\begin{figure}[t]
	\centering
	\begin{minipage}{0.48\textwidth}
		\includegraphics[width=0.9\linewidth]{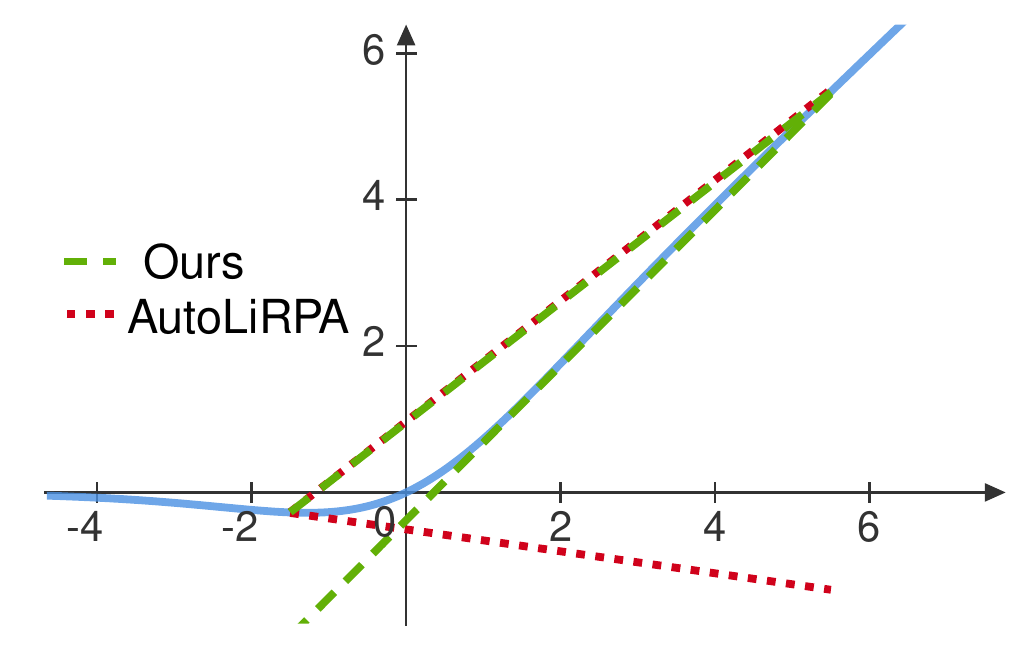}
		\caption{Bounds computed by~\Name{} and~\autolipra{} for $ swish(x)$, $
			x \in [-1.5, 5.5] $.\label{fig:boundcompare2}}
	\end{minipage}%
	\begin{minipage}{0.48\textwidth}
		\centering
		\scalebox{1.0}{
			\begin{tikzpicture}
    \begin{axis}[
        height = .85\textwidth,
        width = \textwidth,
        axis on top = true,
        axis x line = bottom,
        axis y line = left,
        x axis line style = -,
        y axis line style = -,
        tick align = outside,
        every tick/.append style = {
            black,
            thin,
            font=\tiny
        },
        ymin = 0,
        ymax = 1.1,
		xmin=-5,
		xmax=5,
        xlabel = $x_1$,
    ]
        \addplot[
            blue,
            domain = -5:5,
            samples = 100,
			postaction={
        decoration={
          markings,
          mark=between positions 0.38 and .9 step 0.15
               with { \fill circle[radius=1.5pt]; },
        },
        decorate,
      },
        ]
            {1/(1+exp(-x))};

       \addplot[
			name path=f,
            dkgreen,
            domain = -1:3.5,
            samples = 200,
        ]
            {0.649906 + 0.104561*x};

			\node[label={{\textcolor{red}{min shift}}},inner sep=1pt] (source)
			at (axis cs:0.5,.9) {};

		       \node (destination) at (axis
		       cs:2.005,0.881321){};

		\draw[->,>=stealth,red](source.south) to [out=270,in=180] (destination);
	    \addplot[
					thick,
            red,
            domain = 2.005:2.08,
            samples = 200,
        ]
            {0.881321 + -(0.02/0.104561)*(x - 2.005)};

		\path[name path=axis] (axis cs:-1,0) -- (axis cs:3.5,0);
		\addplot [
        thick,
        color=blue,
        fill=blue,
        fill opacity=0.05
    ]
	fill between[
        of=f and axis,
        soft clip={domain=-1:3.5},
    ];

		\node at (axis cs:2.0,0.2)
		{$\displaystyle\int_{-1}^{3.5} c_1^ux_1 +
		c_2^u \, dx$};
    \end{axis}
\end{tikzpicture}
		}
		\caption{Candidate plane synthesis.\label{fig:running_example}}
	\end{minipage}
\end{figure}
Informally, we say a bound is \textit{tight} if the volume below the
upper bound is minimized, and volume below the lower bound is maximized.
Prior work~\cite{zhang2018efficient,SinghGPV19,ko2019popqorn} has
found this to be a good heuristic\footnote{We also experimented with minimizing
the volume between the linear bound and the activation function, which gave
almost identical results.}.
%
Formally, volume is defined as the following integral:
$
\int_{l_1}^{u_1} \dots \int_{l_d}^{u_d} \sum_{i=1}^{d}
c^u_ix_i + c^u_{d+1} \; dx_1 \dots dx_d
$
which, for the upper bound, should be minimized subject to
Equation~\ref{eq:generalsound}. This
integral has the following closed-form solution:
\begin{equation} \label{eq:vol}
\sum_{i=0}^{d} \left[
\frac{1}{2}c_i \times \prod_{j=0}^{d}
\left(
u_i^{1 + \mathbf{1}_{i=j}} - l_i^{1 + \mathbf{1}_{i=j}}
\right)
\right] +
c_{d+1} * \prod_{i=0}^d (u_i - l_i)
\end{equation}
where $ \mathbf{1}_{i=j} $ is the (pseudo Boolean) indicator function that
returns $ 1 $ when its predicate is true.
We omit the proof, but note that the above expression can be
derived inductively on $ d $. Also note that, since each $ l_i, u_i $ are
concrete, the above expression is linear in terms of the coefficients, which
will be advantageous in our approach below.

While recent approaches in solving non-linear optimization
problems~\cite{kong2018delta,chabert2009contractor} could directly minimize
Equation~\ref{eq:vol}
subject to Equation~\ref{eq:generalsound} in one step, we find the runtime to
be very slow. Instead, we adopt a two-step approach that
first uses efficient procedures for computing candidate coefficients that are
almost sound (explained in this section), and second, only calls an SMT
solver when necessary to verify Equation~\ref{eq:generalsound} (explained in
the next
section).
We illustrate the approach on a concrete example.

\subsection{Synthesizing Candidate Bounds}

The first step in our approach computes candidate coefficients for the linear
bound. In this step we focus on satisfying the tightness requirement, while
making a best effort for soundness. We draw inspiration from prior
work~\cite{ryou2021scalable,balunovic2019certifying} that leverages sampling to
estimate the curvature of a particular function, and then uses a linear
programming (LP) solver to compute a plane that is sound. However, unlike prior
work which targeted a fixed function, we target arbitrary (activation)
functions, and thus these are special cases of our approach.


The constraints of the LP are
determined by a set of sample points $ S \subset \mathbb{R}^d $. For the upper
bound, we
minimize Equation~\ref{eq:vol}, subject to the constraint that the linear bound
is above $ \sigma(\cdot) $ at the points in $ S $. Using $ \mathbf{s}_i $ to
refer to the $ i^{th} $ element of the vector $ \mathbf{s} \in S $, the linear
program we solve is:
\begin{equation}\label{eq:lp}
\begin{gathered}
	\text{minimize}\;\text{Equation}\;(\ref{eq:vol}) ~~
	\text{subject to} \bigwedge_{\mathbf{s} \in S} c_1\mathbf{s}_1 +
	c_2\mathbf{s}_2  + \dots + c_{d+1} \geq \sigma(\mathbf{s})
\end{gathered}
\end{equation}
We generate $ S $ by sampling uniformly-spaced points over the input intervals.

\paragraph{Example}

We demonstrate our approach on the running example illustrated in
Fig.~\ref{fig:running_example}. For the example, let
$ \sigma(x_1) = \frac{1}{1 + e^{-x_1}} $ (the sigmoid function, shown as the
blue curve), where $ x_1 \in [-1, 3.5] $. We focus only on the upper bound, but
the lower bound is computed analogously.

Plugging in the variables into
Equation~\ref{eq:vol}, the objective of the LP that we minimize is:
$
\displaystyle\int_{-1}^{3.5} c_1^ux_1 + c_2^u \,dx_1 = 6.625c_1^u + 4.5c_2^u
$
which is shown as the shaded region in Fig.~\ref{fig:running_example}.

We
sample the points  $ S = \{-1, 0.25, 1.5, 2.75\} $, resulting in the following
four constraints:
$ -c_1 + c_2 \geq \sigma(-1) \wedge 0.25c_1 + c_2 \geq \sigma(0.25) \wedge
1.5c_1 + c_s \geq \sigma(1.5) \wedge 2.75c_1 + c_2 \geq \sigma(2.75) $.	Solving
the LP program results in $ c_1 = 0.104, c_2 = 0.649 $, which is illustrated by the
green line in Fig.~\ref{fig:running_example}.

\section{Making the Bound Sound}
\label{sec:method-2}

In this section, we present our method for obtaining soundness because the
candidate bounds synthesized in the previous section may not be
sound. Here, we focus only on making the upper bound sound, but note the
procedure for the lower bound is similar.

\subsection{Problem Statement and Challenges}
We are given the activation function $ \sigma(\cdot) $, the input intervals $
x_i \in [l_i, u_i] $, and the candidate coefficients $ c_1, c_2,
\dots,c_{d+1} $. The goal is to compute an upward shift, if needed, to make the upper bound sound. First,
we define the violation of the upper bound as:
\begin{equation}
	v(x_1, x_2,\dots,x_d) := c^u_1x_1 + c^u_2x_2 + \dots + c^u_{d+1} -
	\sigma(x_1, x_2, \dots,x_d)
\end{equation}
A negative value indicates the upper bound is not sound. We then need to compute a
lower bound on $ v(\cdot) $, which we term $ v_l $. Then the equation we pass
to the verifier is:
\begin{equation} \label{eq:makesound}
\begin{gathered}
\forall x_1 \in [l_1, u_1], x_2 \in [l_2, u_2], \dots, x_d\in[l_d, u_d]\\
v(x_1, x_2, \dots, x_d) + (-v_l) \geq 0
\end{gathered}
\end{equation}
Expanding $ v(\cdot) $ with its definition in the above equation results in the
soundness definition of Equation~\ref{eq:generalsound}. Thus, if the verifier
proves Equation~\ref{eq:makesound}, then shifting the upper
bound upward by $ -v_l $ ensures its soundness.
For our running example, the quantity $ v_l $ is shown by the red line in
Fig.~\ref{fig:running_example}.

This problem is non-trivial because finding a solution for $ v_l $ requires a
search for a sound global minimum/maximum of a function involving $
\sigma(\cdot) $, which may be highly non-linear.
State-of-the-art  SMT solvers such as  Z3 do not support all non-linear
operations, and furthermore, since we assume arbitrary $ \sigma(\cdot) $, the
problem may even be (computationally) undecidable.

\subsection{Verifying the Bound}
We first assume we have a candidate (possibly unsound) $ v_l $, and explain our
verification method.
To ensure decidability and tractability, we
leverage the \textit{$ \delta $-decision procedure} implemented
by~\dReal{}~\cite{gao2013dreal}. To the best of our knowledge this is is the only
framework that is decidable for all computable functions.

In this context, instead of verifying Equation~\ref{eq:makesound}, the formula
is first negated thus changing it into an existentially quantified one, and
then applying a $ \delta $\textit{-relaxation}. Formally, the formula~\dReal{}
attempts to solve is:
\begin{equation} \label{eq:relaxed_makesound}
\begin{gathered}
\exists x_1 \in [l_1, u_1], x_2 \in [l_2, u_2], \dots, x_d \in [l_d, u_d]\\
v(x_1, x_2, \dots) + (-v_l) \leq \delta
\end{gathered}
\end{equation}
where $ \delta $ is a small constant (e.g. $ 10^{-5} $), which we explain in a
moment. The
above is formulated such that Equation~\ref{eq:makesound} holds if (but not
only if) there does \textit{not} exist a solution to
Equation~\ref{eq:relaxed_makesound}.

Internally,~\dReal{} performs interval constraint propagation (ICP) on the
left-hand
side of Equation~\ref{eq:relaxed_makesound} over the
intervals defined by each $ [l_i, u_i] $ to compute an upper bound, and
compares this upper bound with $ \delta $. If the upper bound is less than $
\delta $, then no solution exists (i.e., Equation~\ref{eq:relaxed_makesound} is
unsatisfiable, and we have proven the original Equation~\ref{eq:makesound} holds). Otherwise
a solution \textit{may} exist. In this case,~\dReal{} iteratively partitions
the input space defined by the $ [l_i, u_i ] $ and repeats this process on
each partition separately.

\dReal{} stops partitioning either when
it proves all partitions do not have solutions
, or when a partition whose intervals all
have width less than some $ \epsilon $ is found. Here, $ \epsilon $ is
proportional to $
\delta $ (i.e., smaller $ \delta $ means smaller $ \epsilon $). In the latter
case,~\dReal{} returns this partition as a ``solution''.

While Equation~\ref{eq:makesound} holds if there does not exist a solution to
Equation~\ref{eq:relaxed_makesound}, the converse does not hold true both
because of the error inherent in ICP, and because
we ``relaxed'' the right-hand side of  Equation~\ref{eq:relaxed_makesound}.
This means that $ \delta $ controls the
\textit{precision} of the analysis. $ \delta $ controls both the size of the
false solution space, and determines how many times we will sub-divide the
input space before giving up on proving Equation~\ref{eq:relaxed_makesound} to
be unsatisfiable.

Practically, this has two implications for our approach. The first one is that our
approach naturally inherits a degree of looseness in the linear bounds defined
by $ \delta $. Specifically, we must shift our plane upward by $
\delta $ in addition to the true $ v_l $, so that~\dReal{} can verify the
bound. The second is that we have to make a trade-off between computation and precision. While
smaller $ \delta $ will allow us to verify a tighter bound, it generally will
also mean a longer verification time.
In our experiments, we find that $ \delta = 10^{-7} $ gives tight bounds at an
acceptable runtime, though we may be able to achieve a shorter  runtime with a
larger $\delta$.

\subsection{Computing $ v_l $}
Now that we have defined how we can verify a candidate bound, we explain our
approach for computing $ v_l $. The implementation is outlined in
Algorithm~\ref{alg:viol}. Since failed calls to the verifier can be
expensive, at lines 1-2, we first use a relatively cheap (and unsound) local
optimization procedure to estimate the true $ v_l $.
While local optimization may get stuck
in local minima, neural network activation functions typically do not have many
local minima, so neither will $ v(\cdot) $.
We use L-BFGS-B~\cite{byrd1995limited}, the bounded version of L-BFGS, to
perform the optimization. At a high-level, L-BFGS-B takes as input $ v(\cdot)
$, the input bounds $ x_i \in [l_i, u_i] $, and an initial guess $ \textbf{g}
\in
\mathbb{R}^d $ at the location of the local minimum.
It then uses the
Jacobian matrix (i.e., derivatives) of $ v(\cdot) $ to iteratively move towards
the local minimum (the Jacobian can be estimated using the finite differences
method or provided explicitly -- we use Mathematica~\cite{Mathematica} to
obtain it).
We find that sampling points uniformly in $ v(\cdot) $ can usually find a good
$ \textbf{g} $, and thus L-BFGS-B often converges in a small number of
iterations.
L-BFGS-B typically produces an estimate within $ 10^{-8} $ of the true value.
To account for estimation error we add an additional $ 10^{-6} $, plus $ 2
\times \delta $ to account for the $ \delta $-relaxation (line 3).
Finally, we
iteratively decrease $ v_l $ by a small amount ($ 10^{-6} $)
until~\dReal{} verifies it (lines 4-9).

Going back to our motivating example, we would estimate $ v_l $ with a local
minimizer, and then use \dReal{} to verify the following:
\begin{gather*}
\forall x_1 \in [-1, 3.5] ~. \; \sigma(x_1) \leq c_1^ux_1 + c_2^u + (-v_l) + 2
\times \delta + 10^{-6}
\end{gather*}
If verification fails, we iteratively decrease the value of $ v_l $ by $
10^{-6} $, and call \dReal{} until the bound is verified. The final
value of $ c_1^ux_1 + c_2^u + (-v_l) + 2 \times \delta + 10^{-6} $ is the final
sound upper bound.

\subsection{On the Correctness and Generality of~\Name{}}
The full~\Name{} procedure is shown in Algorithm~\ref{alg:synth}. The
correctness (i.e. soundness) of the synthesized bounds is guaranteed if the $
v_l $ returned by Algorithm~\ref{alg:viol} is a true lower bound on $ v(\cdot)
$. Since Algorithm~\ref{alg:viol} does not return until~\dReal{} verifies $ v_l
$ at line 6, the correctness is guaranteed.

Both our procedure in Section~\ref{sec:method-1} and L-BFGS-B require only
black-box access to $ \sigma(\cdot) $, so the only potential limit to the
arbitrariness of our approach lies in what elementary operations are supported
by~\dReal{}. During our investigation, we did not find activations that use
operations unsupported by~\dReal{}, however if an unsupported operation is
encountered, one would only need to define an \textit{interval
extension}~\cite{moore2009introduction} for the operation, which can be done
for any computable function.

\begin{algorithm}[t]
	\SetAlgoLined
	\KwIn{Activation $ \sigma(x_1, x_2, \dots) $, Candidate Coefficients
	$c_1^u, c_2^u, \dots, c_{d+1}^u$, $ ~~~~ $Input Bounds
		$x_1 \in [l_1, u_1], x_2 \in [l_2, u_2],\dots$, Jacobian $\nabla v$
		(optional)}
	\KwOut{Lower Bound on Violation $v_l$}
	$\textbf{g} \gets $ sample points on $v(x_1,x_2,\dots)$ and take minimum\;
	$v_l \gets \textbf{L-BFGS-B}( v(x_1,x_2,\dots),
	x_1 \in [l_1, u_1], x_2 \in [l_2, u_2], \dots, \textbf{g}, \nabla v )$ \;
	$ v_l \gets v_l - 10^{-6} - 2\delta$\;
	\While{$\textbf{True}$}{
		// Call dReal \\
		\If{$ \text{Equation~\ref{eq:generalsound} holds} $}
		{
			\Return{} $v_l$;
		}
		$v_l \gets v_l - 10^{-6}$;
	}
	\caption{BoundViolation\label{alg:viol}}
\end{algorithm}
\begin{algorithm}[t]
	\SetAlgoLined
	\KwIn{Activation $\sigma(x_1, x_2, \dots)$, Input Bounds $x_1 \in [l_1,
	u_1],
		x_2 \in [l_2, u_2], \dots$, Jacobian $\nabla v$ (optional)}
	\KwOut{Sound Coefficients $c_1^u, c_2^u, \dots, c_{d+1}^u$}
	$c_1^u, c_2^u, \dots, c_{d+1}^u \gets  \text{Sampling and LP procedure on }
	\sigma(x) \text{ over Input Bounds}$\;
	$v_l \gets \text{BoundViolation}(c_1^u, c_2^u, \dots, c_{d+1}^u, x_1 \in
	[l_1, u_1], x_2 \in [l_2, u_2], \dots, \nabla v)$\;
	$c_{d+1}^u \gets c_{d+1}^u + (-v_l)$\;
	\Return{} $c_1^u, c_2^u, \dots, c_{d+1}^u $\;
	\caption{SynthesizeUpperBoundCoefficients\label{alg:synth}}
\end{algorithm}

\section{Evaluation}
\label{sec:experiment}

We have implemented our method in a module called~\Name{}, and
integrated it into the \autolipra{} neural network verification framework~\cite{autolipra}.
A user instantiates \Name{} with a definition of an activation
function, which results in an executable software module capable of
computing the sound linear lower and upper bounds for the activation
function over a given input region.
\Name{} uses Gurobi~\cite{gurobi} to solve the LP problem described in
Section~\ref{sec:method-1}, and \dReal{}~\cite{gao2013dreal} as the verifier
described in~\ref{sec:method-2}.
In total, \Name{} is implemented in about 1200 lines of Python
code.

\subsection{Benchmarks}

\paragraph{Neural Networks}
Our benchmarks are nine deep neural networks trained on the three
different datasets shown below. In the following, a neuron is a node in the
neural network where a linear bound must be computed, and thus the neuron counts
indicate the number of calls to~\Name{} that must be made.
\begin{itemize}
\item
\textbf{MNIST:}
MNIST is a dataset of hand-written integers labeled with the
corresponding integer in the image. The images have 28x28 pixels, with
each pixel taking a gray-scale value between 0 to 255. We trained
three variants of a 4-layer CNN (convolutional neural network). Each takes as
input a 28x28 = 784-dimensional input vector and outputs 10 scores, one for
each
class. In total, each network has 2,608 neurons -- 1568, 784, and 256 in
the first, second, and third layers, respectively.

\item
\textbf{CIFAR:}
CIFAR is a dataset of RGB images from 10 different classes. The images
have 32x32 pixels, with each pixel having an R, G, and B value in the
range 0 to 255. We trained three variants of a 5-layer CNN. Each takes a
32x32x3 = 3072-dimensional input vector and outputs 10 scores, one for each
class. In total, each network has 5376 neurons, 2048, 2048, 1024, and 256
neurons in the first, second, third, and fourth layers, respectively.

\item
\textbf{SST-2:}
The Stanford Sentiment Treebank (SST) dataset consists of sentences
taken from movie reviews that are human annotated with either positive
or negative, indicating the sentiment expressed in the sentence. We
trained three different variants of the standard LSTM architecture. These
networks take as input a sequence 64-dimensional word embeddings and
output 2 scores, one for positive and one for negative. Each network has a
hidden size of 64, which works out to 384 neurons per input in the input
sequence.
\end{itemize}

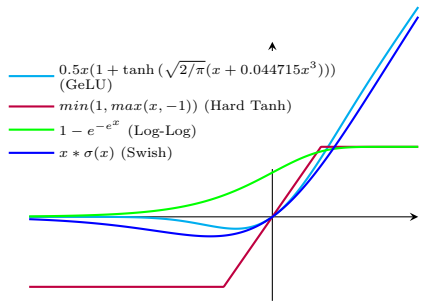
\begin{wrapfigure}{R}{0.55\textwidth}
	\centering
	\begin{tikzpicture}
        \begin{axis}[
            xmin = -5, xmax = 3,
            ymin = -1.2, ymax = 2.5,
            xtick distance = 10,
            ytick distance = 10,
            width = \linewidth,
            height = 0.75\linewidth,
            xticklabel=\empty,yticklabel=\empty,
            minor tick num=0,
            axis lines = middle,
            legend cell align = {left},
			set layers=standard,
	        legend to name=grouplegend,
	        legend entries={{$ 0.5 x ( 1 + \tanh{( \sqrt{2 / \pi } (x + 0.044715
	        x ^{3} ) )} ) $ \\ (GeLU)},
	        	{$ min(1, max(x, -1)) $ (Hard Tanh)},
	        	{$ 1 - e^{-e^{x}} $ (Log-Log)},
	        	{$ x * \sigma(x) $ (Swish)},},
        	legend style={nodes={scale=0.75, transform shape},font=\scriptsize,
        	draw=none,fill=white,align=left},
        ]
            \addplot[actfunc, cyan] {0.5*x * ( 1 + tanh(sqrt(2/pi) * (x+
            	0.044715 * (x ^3) ) ) )};
            \addplot[actfunc, purple] {min(max(-1, x), 1)};
            \addplot[actfunc, green] {1 - exp(-exp(x))};
            \addplot[actfunc, blue] {x * (1 / (1 + exp(-x)))};

			\coordinate (leg) at (rel axis cs:-0.1,1);

        \end{axis}
        \node[anchor= north west] at
        (leg){\pgfplotslegendfromname{grouplegend}};
\end{tikzpicture}
	\caption{Nonlinear activation functions.\label{fig:actfuncs}}
\end{wrapfigure}
\paragraph{Activation Functions}
We experimented with the four activation functions as shown in
Fig.~\ref{fig:actfuncs}.
\emph{GELU} and \emph{Swish} were recently proposed
alternatives to the standard ReLU
function due to their desirable
theoretical properties~\cite{hendrycks2016gaussian} such as reduced
overfitting~\cite{singla2021low}, and they have seen use in
OpenAI's GPT~\cite{radford2018improving} and very deep feed forward
networks~\cite{ramachandran2017searching}.
Similarly, \emph{Hard-Tanh} is an optimized version of the common
$\tanh{}$ function, while the \emph{Log-Log}
function~\cite{gomes2008complementary} is a sigmoid-like function
used in forecasting.

\paragraph{The Verification Problem}
The verification problem we consider is to certify that an input is robust to
bounded perturbations of magnitude $ \epsilon $, where $\epsilon$ is a small number. \textit{Certifying} means
proving that the classification result of the neural network does not change in the presence of
perturbations. We focus on $ l_{\infty} $ robustness, where we take an input $
\mathbf{x} \in \mathbb{R}^n $ and allow a bounded perturbation of $ +/-
\epsilon $ to each element in $ \mathbf{x} $. For each network, we take 100
random test inputs, filter out those that are incorrectly classified, apply an
$ \epsilon $ bounded perturbation to the correctly classified inputs, and then
attempt to prove the classification remains correct. We choose $ \epsilon $
values common in prior work. For MNIST networks, in particular, we choose $ \epsilon = 8/255
$. For CIFAR networks, we  choose $ \epsilon = 1/255 $. For SST-2 networks, we
choose $ \epsilon = 0.04 $, and we only apply it to the first word embedding in
the input sequence.

\subsection{Experimental Results}

Our experiments were designed to answer the following two questions:
(1) How do~\Name{}'s linear bounds compare with handcrafted bounds?
(2) How does the runtime of~\Name{} compare to state-of-the-art linear
bounding techniques?
To answer these questions, we compare the effectiveness of~\Name{}'s
linear bounds with the state-of-the-art linear bounding technique
implemented in~\autolipra{}. To the best of our knowledge this is the only tool
that can handle the activation functions we use in our benchmarks.
As another comparison point, we also compare
with~\popqorn{}, a state-of-the-art linear bounding technique for LSTM
networks.~\popqorn{} tackles the challenge of computing tight linear bounds for
$ sigmoid(x) \times tanh(y) $ and $ x \times sigmoid(y) $ using an expensive
gradient descent based approach, and thus makes a good comparison point for
runtime and accuracy.
Our experiments were conducted on a computer with an Intel 2.6 GHz i7-6700
8-core CPU and 32GB RAM.
Both~\autolipra{} and~\Name{} are engineered to bound individual neurons in
parallel. We configure each method to use up to 6 threads.

\paragraph{Overall Comparison}

\begin{table}[t]
	\centering
	\caption{Comparing certified accuracy and run time of \Name{} and
	\autolipra{}.}
	\label{tbl:1}
	\scalebox{0.85}{
		\begin{tabular}{|ll|c|r|c|r|}\hline
			& \multicolumn{1}{c|}{\multirow{2}{*}{Network Architecture}}
			& \multicolumn{2}{l|}{\autolipra{}~\cite{autolipra}}
			& \multicolumn{2}{l|}{Our Method	(new)}          \\ \cline{3-6}
			& \multicolumn{1}{c|}{} & \% certified       & time (s)
			& \multicolumn{1}{l|}{\% certified} & time(s)		\\ \hline\hline

			\multicolumn{1}{|l|}{MNIST} & 4-Layer CNN with Swish   & 0.34
			&    15 & 0.76   &   796   \\ \cline{2-6}
			\multicolumn{1}{|l|}{}      & 4-Layer CNN with Gelu    & 0.01   &
			359 & 0.72   &   814   \\ \cline{2-6}
			\multicolumn{1}{|l|}{}      & 4-Layer CNN with Log Log & 0.00
			&    38 & 0.24   &   867   \\ \hline\hline
			\multicolumn{1}{|l|}{CIFAR} & 5-Layer CNN with Swish   & 0.03
			&    69 & 0.35   & 1,077   \\ \cline{2-6}
			\multicolumn{1}{|l|}{}      & 5-Layer CNN with Gelu    & 0.00   &
			1,217 & 0.31   & 1,163   \\ \cline{2-6}
			\multicolumn{1}{|l|}{}      & 5-Layer CNN with Log Log & 0.59
			&    98 & 0.69   &   717   \\ \hline\hline
			\multicolumn{1}{|l|}{SST-2} & LSTM with sig tanh       & 0.93
			&    37 & 0.91   & 1,074   \\ \cline{2-6}
			\multicolumn{1}{|l|}{}      & LSTM with hard tanh      & -
			& -      & 0.64   &    2300     \\ \cline{2-6}
			\multicolumn{1}{|l|}{}      & LSTM with log  log       & 0.16   &
			1,072 & 0.82   & 2,859   \\ \hline
		\end{tabular}
	}
\end{table}

\begin{table}[t]
	\centering
	\caption{Comparing certified accuracy and run time of~\Name{} and
		\popqorn{}.}
	\label{tbl:2}
	\scalebox{0.85}{
		\begin{tabular}{|ll|c|r|c|r|}\hline
			& \multicolumn{1}{c|}{\multirow{2}{*}{Network Architecture}}
			& \multicolumn{2}{l|}{\popqorn{}~\cite{ko2019popqorn}}
			& \multicolumn{2}{l|}{Our Method	(new)}          \\ \cline{3-6}
			& \multicolumn{1}{c|}{} & \% certified       & time (s)
			& \multicolumn{1}{l|}{\% certified} & time(s)		\\ \hline\hline

			\multicolumn{1}{|l|}{SST-2} & LSTM with sig tanh       & 0.93
			&    1517 & 0.90   & 1,074  \\ \hline
		\end{tabular}
	}
\end{table}

First, we compare the overall performance of our new method and the
default linear bounding technique in~\autolipra{}.  The results are shown
in Table~\ref{tbl:1}.  Here, Columns~1-2 show the name of the dataset
and the type of neural networks.  Columns~3-4 show the results of the
default~\autolipra{}, including the percentage of inputs certified and
the analysis time in seconds.  Similarly, Columns~5-6 show the results
of our new method.

The results in Table~\ref{tbl:1} show that, in terms of the analysis time,
our method is slower, primarily due to the use of constraint solvers (namely~\dReal{} and
the LP solver) but overall, the analysis speed is
comparable to~\autolipra{}.
However, in terms of accuracy, our method significantly
outperforms~\autolipra{}.  In almost all
cases, our method was able to certify a much higher percentage of the inputs.
For example,~\Name{} more than quadruples the certified robustness of the
\emph{LSTM with log log} benchmark, and handles very well the relatively
complex GeLU function.
As for \emph{SST-2: LSTM with hard tanh},~\autolipra{} does not support the
general $ max(x, y) $ operation, so a comparison is not possible without
significant engineering work.

The only exception to the improvement is \emph{SST-2: LSTM with sig tanh},
for which the results are similar (.93 versus .91).
In this case, there is likely little to be gained over the default, decomposition-based approach of \autolipra{} in terms of tightness because
the inputs to $ sigmoid(x) \times tanh(y) $ and $ x \times sigmoid(y) $ are not
related, i.e., $ x $ and $ y
$ are two separate variables. This is in contrast to, e.g., $ swish(x) = x
\times sigmoid(x) $, where the left-hand side and right-hand side of the
multiplication \textit{are} related.

In Table~\ref{tbl:2}, we show a comparison between~\Name{} and~\popqorn{}. The
result shows that our approach achieves similar certified robustness and
runtime, even though~\popqorn{} was designed to specifically target this particular type of  LSTM architecture,
while~\Name{} is entirely generic.

\paragraph{Detailed Comparison}

\begin{figure}[t]
	\centering
	\begin{minipage}{.42\textwidth}
		\centering
		\includegraphics[width=\linewidth]{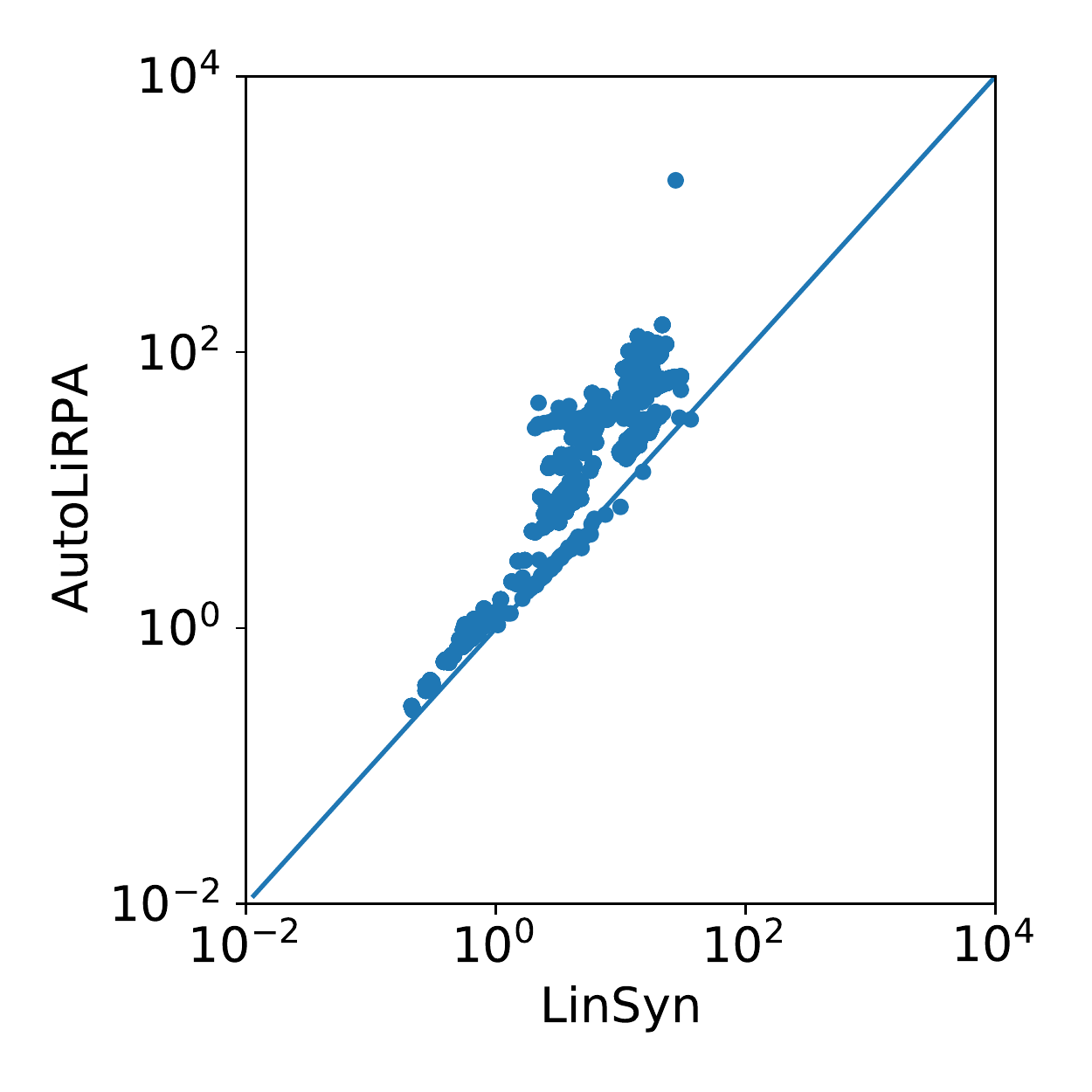}
		\caption{Scatter plot comparing the final output interval width of
		\Name{} and~\autolipra{}.}
		\label{fig:scatter-plot}
	\end{minipage}\hspace{24pt}%
	\begin{minipage}{.5\textwidth}
		\centering
		\includegraphics[width=\linewidth]{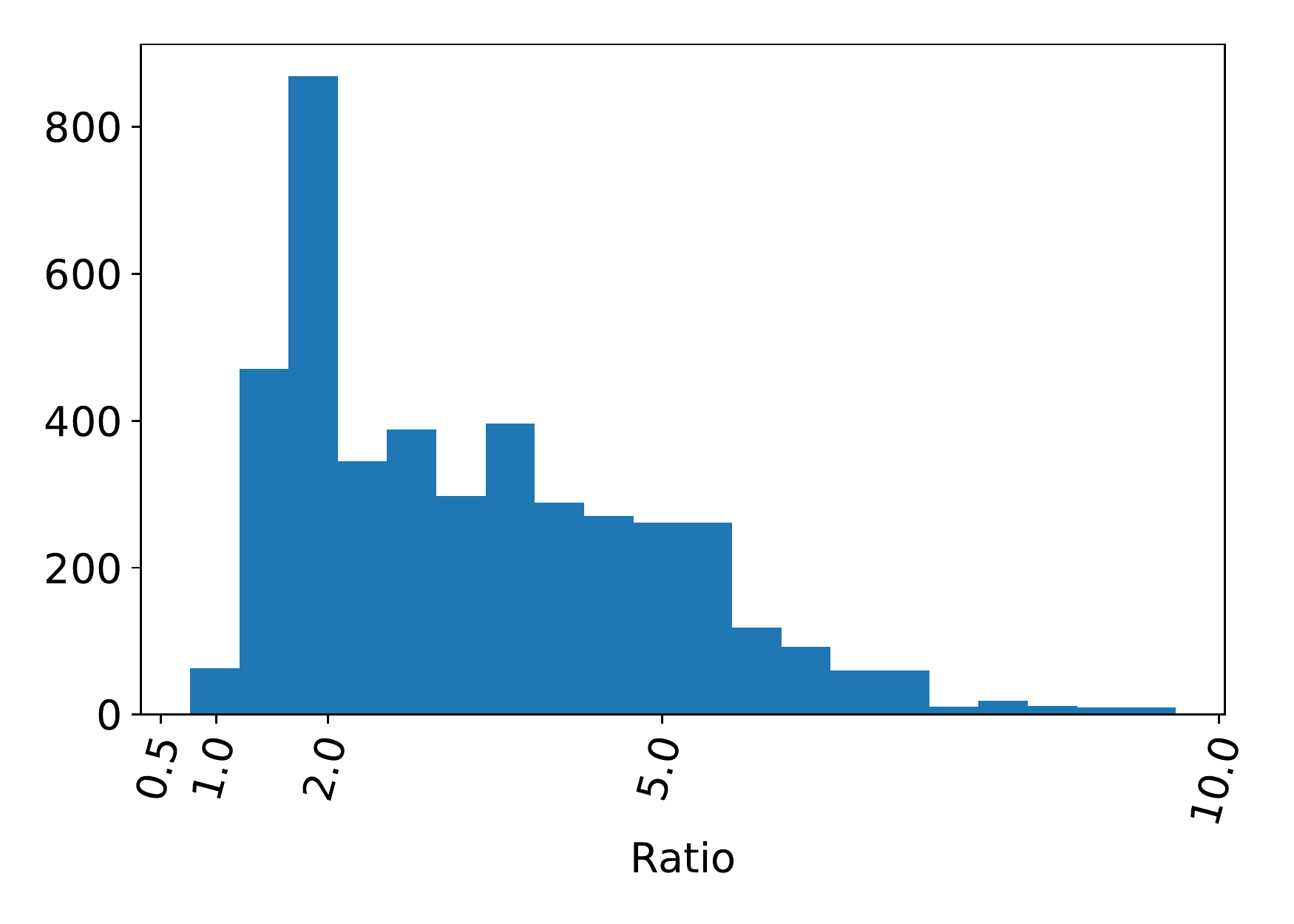}

		\caption{Histogram of width ratios between \autolipra{} and \Name{}. Ratio reported as $
		\frac{\autolipra{}}{\Name{}} $.}
		\label{fig:ratio-hist}
	\end{minipage}
\end{figure}


Next, we perform a more in depth comparison of accuracy by comparing the widths
of the final output neuron's intervals that are computed by~\autolipra{}
and~\Name{}. The results are shown in the scatter plot in
Fig.~\ref{fig:scatter-plot} and the histogram in Fig.~\ref{fig:ratio-hist}.
Each point in the scatter plot represents a single output neuron $ x_i
$ for a single verification problem. The $ x $-axis is the width of the
interval of the output neuron $ x_i $ (i.e. $ u_i - l_i $) computed by~\Name{},
and the $ y $-axis is the width computed by~\autolipra{}. A point above the
diagonal line
indicates that~\Name{} computed a tighter (smaller) final output interval.
In the histogram, we further illustrate the accuracy gain as the width ratio,
measured as $ \frac{\autolipra{}}{\Name{}} $.
%
Overall, the results show that~\Name{} is more accurate in nearly all cases,
and~\Name{} often produces final output bounds 2-5X tighter than~\autolipra{}.

\section{Related Work}
\label{sec:related}

\textit{Linear Bound-based Neural Network Verification}
There is a large body of work on using linear-bounding
techniques~\cite{SinghGPV19,zhang2018efficient,shi2020robustness,boopathy2019cnn,WengZCSHDBD18,paulsen2020reludiff,paulsen2020neurodiff,wu2021tightening,mohammadinejad2020diffrnn}
and other abstract domains such as concrete intervals, symbolic
intervals~\cite{WangPWYJ18}, and Zonotopes~\cite{GehrMDTCV18},
for the purpose of neural network verification.
All of these can be thought of as leveraging restricted versions of the
polyhedral abstract domain~\cite{CousotH78,CousotC77}.
%
To the best
of our knowledge, these approaches are the most scalable (in terms of network
size) due to the use of approximations, but this also means they are less
accurate than exact approaches. In addition, all these approaches have the
limitation that they depend on bounds that are hand-crafted by an expert.
%


\textit{SMT solver-based Neural Network Verification}
There is also a large body of work on using exact constraint solving for neural
network verification. Early works include solvers specifically designed for
neural networks, such as Reluplex and
Marabou~\cite{KatzBDJK17,KatzHIJLLSTWZDK19} and others~\cite{DvijothamSGMK18},
and leveraging existing
solvers~\cite{Ehlers17,HuangKWW17,BastaniILVNC16,HuangKWW17,baluta2019quantitative,tjeng2019evaluating,hu2020reach}.
 While more accurate, the reliance on an SMT solver typically limits their
 scalability. More recent work
often uses solvers to refine the bounds computed by linear
bounding~\cite{Singh2019krelu,SinghGPV19iclr,WangPWYJ18nips,tran2019star,tran2020verification}.
Since the solvers leveraged in these approaches usually involve linear
constraint solving techniques, they are usually only applicable to piece-wise
linear activation functions such as ReLU and Max/Min-pooling.


%

%

\section{Conclusions}
\label{sec:conclusion}

We have presented~\Name{}, a method for synthesizing linear bounds for
arbitrary activation functions.
%
%
The key advantage of~\Name{} is that it can handle complex
activation functions, such as Swish, GELU, and Log Log as a whole, allowing it
to synthesize much tighter linear bounds than existing tools.
Our experimental
results show this increased tightness leads to drastically increased certified
robustness, and tighter final output bounds.



\newpage\clearpage
\bibliographystyle{splncs04}
\bibliography{convexsyn}
\vfill

{\small\medskip\noindent{\bf Open Access} This chapter is licensed under the
terms of the Creative Commons\break Attribution 4.0 International License
(\url{http://creativecommons.org/licenses/by/4.0/}), which permits use,
sharing, adaptation, distribution and reproduction in any medium or format, as
long as you give appropriate credit to the original author(s) and the source,
provide a link to the Creative Commons license and indicate if changes were
made.}

{\small \spaceskip .28em plus .1em minus .1em The images or other third party
material in this chapter are included in the chapter's Creative Commons
license, unless indicated otherwise in a credit line to the material.~If
material is not included in the chapter's Creative Commons license and your
intended\break use is not permitted by statutory regulation or exceeds the
permitted use, you will need to obtain permission directly from the copyright
holder.}

\medskip\noindent\includegraphics{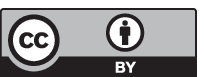}

\end{document}